\documentclass[letterpaper,10pt,conference]{ieeeconf}
\IEEEoverridecommandlockouts
\usepackage[font=small,labelfont=bf]{caption}
\usepackage{ieee_robotics}

\newcommand{\method}{Vi-TacMan\xspace}
\newcommand{\transpose}{\mathsf{T}}

\acrodef{dof}[DoF]{degree of freedom}
\acrodefplural{dof}[DoFs]{degrees of freedom}
\acrodef{map}[MAP]{maximum a posteriori}
\acrodef{pdf}[PDF]{probability density function}
\acrodef{vmf}[vMF]{von Mises-Fisher}
\acrodef{mle}[MLE]{Maximum Likelihood Estimation}
\acrodef{smom}[SMOM]{Stein's Method of Moments}

\usepackage{orcidlink}
\usepackage[detect-weight,mode=text]{siunitx}
\hypersetup{
  colorlinks=true,
}

\title{\LARGE \bf \method: Articulated Object Manipulation via Vision and Touch}

\author{%
Leiyao Cui$^{\orcidlink{0009-0009-4925-6983}\,1,2,3,4,6}$\,\,%
Zihang Zhao$^{\orcidlink{0000-0003-3215-7152}\,3,4,5,6,7,\dagger}$\,\,%
Sirui Xie$^{\orcidlink{0009-0003-9379-2122}\,3}$\,\,%
Wenhuan Zhang$^{\orcidlink{0009-0003-7925-124X}\,3}$\,\,%
Zhi Han$^{\orcidlink{0000-0002-8039-6679}\,1,2,\dagger}$\,\,%
Yixin Zhu$^{\orcidlink{0000-0001-7024-1545}\,4,3,5,6,8,\dagger}$%
\vspace{6pt}%
\\\url{https://vi-tacman.github.io}%
\thanks{L. Cui, Z. Zhao, S. Xie, and W. Zhang contributed equally to this work.
$^\dagger$ Corresponding authors. Emails:
\texttt{zhaozihang@stu.pku.edu.cn}, \texttt{hanzhi@sia.cn}, and
\texttt{yixin.zhu@pku.edu.cn}.
}%
\thanks{
$^1$ Shenyang Institute of Automation, Chinese Academy of Sciences
$^2$ University of Chinese Academy of Sciences
$^3$ Institute for Artificial Intelligence, Peking University
$^4$ School of Psychological and Cognitive Sciences, Peking University
$^5$ State Key Lab of General AI, Peking University
$^6$ Beijing Key Laboratory of Behavior and Mental Health, Peking University
$^7$ LeapZenith AI Research
$^8$ Embodied Intelligence Lab, PKU-Wuhan Institute for Artificial Intelligence%
}%
\thanks{This work is supported in part by the Brain Science and Brain-like Intelligence Technology--National Science and Technology Major Project (2025ZD0219400), the National Natural Science Foundation of China (62376009), the State Key Lab of General AI at Peking University, the PKU-BingJi Joint Laboratory for Artificial Intelligence, the Wuhan Major Scientific and Technological Special Program (2025060902020304), the Hubei Embodied Intelligence Foundation Model Research and Development Program, and the National Comprehensive Experimental Base for Governance of Intelligent Society, Wuhan East Lake High-Tech Development Zone.}
}

\begin{document}

\maketitle
\begin{abstract}
Autonomous manipulation of articulated objects remains a fundamental challenge for robots in human environments.
Vision-based methods can infer hidden kinematics but can yield imprecise estimates on unfamiliar objects. Tactile approaches achieve robust control through contact feedback but require accurate initialization. This suggests a natural synergy: vision for global guidance, touch for local precision.
Yet no framework systematically exploits this complementarity for generalized articulated manipulation.
Here we present \method, which uses vision to propose grasps and coarse directions that seed a tactile controller for precise execution.
By incorporating surface normals as geometric priors and modeling directions via \ac{vmf} distributions, our approach achieves significant gains over baselines (all \(\boldsymbol{p}\)\textless0.0001). Critically, manipulation succeeds without explicit kinematic models---the tactile controller refines coarse visual estimates through real-time contact regulation.
Tests on more than \num{50000} simulated and diverse real-world objects confirm robust cross-category generalization.
This work establishes that coarse visual cues suffice for reliable manipulation when coupled with tactile feedback, offering a scalable paradigm for autonomous systems in unstructured environments.
\end{abstract}

\section{Introduction}

Household robots must master articulated object manipulation to function effectively in human environments, yet face enormous diversity in object appearance, geometry, and kinematics~\cite{xiang2020sapien,liu2022akb,wang2023rearrange,jin2025artvip}. Unlike structured industrial settings where objects are standardized, everyday articulated structures---cabinets, refrigerators, ovens---exhibit vast variability that renders precise a priori modeling impractical~\cite{kemp2007challenges}. This variability poses a fundamental challenge: reliable manipulation requires both accurate localization of interaction points and precise execution of kinematically-constrained motions. The question then becomes: which sensory modality is best suited to address each aspect of this challenge?

Dominant approaches rely on vision to reconstruct object kinematics for manipulation planning~\cite{mo2021where2act,jain2021screwnet,zeng2021visual,eisner2022flowbot3d,mittal2022articulated,yu2024gamma,wang2024rpmart,wang2025adamanip}. Vision's global receptive field makes it well-suited for identifying interaction points across the entire object. However, articulation mechanisms are typically hidden within object interiors, forcing vision systems to infer kinematics from limited surface observations. This inverse problem proves brittle on unfamiliar objects: even state-of-the-art methods trained on large-scale datasets~\cite{xiang2020sapien,liu2022akb,jin2025artvip} produce imprecise kinematic estimates that fail during execution---particularly problematic in safety-critical home environments where reliability is paramount.

\begin{figure}[t!]
    \centering
    \includegraphics[width=\linewidth]{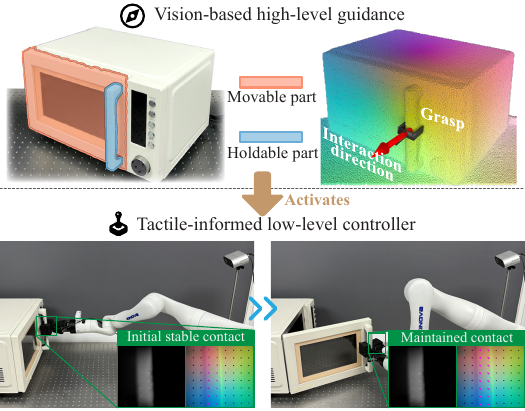}
    \caption{\textbf{Overview of \method.} \method exploits the complementary strengths of vision and touch for manipulating unseen articulated objects. Vision provides global context to propose grasps and estimate coarse interaction directions, which initialize a tactile controller that leverages local contact feedback for precise and robust execution.}
    \label{fig:teaser}
\end{figure}

Recent tactile methods offer an alternative paradigm~\cite{zhao2024tac,zhao2025tacman}: rather than recovering precise kinematics, they maintain successful manipulation through continuous contact regulation. By directly sensing contact geometry, tactile feedback provides rich local information that vision cannot access. Critically, these approaches demonstrate that stable contact feedback enables reliable execution given only coarse initial conditions---a feasible grasp and approximate motion direction. This insight reframes the vision problem: precise kinematic recovery is unnecessary if vision provides sufficient cues to initialize tactile control. The natural division of labor emerges: vision for global, coarse guidance; touch for local, precise execution.

We present \method, a systematic framework exploiting this complementarity. Vision detects movable and holdable parts, proposes grasps, and estimates coarse interaction directions; tactile feedback then refines execution through real-time contact regulation (\cref{fig:teaser}). Three key technical components enable robust generalization to unseen objects. First, we incorporate surface normals as geometric priors for direction estimation, providing physical constraints that significantly improve performance (\(p\)\textless0.0001). Second, recognizing that multiple plausible directions may exist for unfamiliar objects, we model directional uncertainty via \acfp{vmf} on the unit sphere~\cite{mardia2009directional}, enabling principled inference under ambiguity. Third, our detector achieves 0.86 mAP~\cite{lin2014microsoft}, reliably identifying interaction regions even in complex multi-part objects. Together, these components provide the sufficient initialization required by tactile control.

Our contributions are:
\begin{itemize}[leftmargin=*,noitemsep,nolistsep]
\item We present \method, a vision-touch framework where coarse visual guidance activates precise tactile control for articulated manipulation.
\item We develop a robust detection model achieving 0.86 mAP~\cite{lin2014microsoft} that identifies movable and holdable parts in complex multi-component objects.
\item We incorporate surface normals as geometric priors for direction estimation, yielding significant gains over baselines (all \(p\)\textless0.0001).
\item We apply \acf{vmf} distributions to model directional uncertainty on the unit sphere, enabling principled inference under ambiguity.
\item We validate our approach on over \num{50000} simulations and diverse real objects, demonstrating reliable manipulation without explicit kinematic models.
\end{itemize}

The remainder of this paper is organized as follows: \cref{sec:method} presents our systematic approach to articulated object manipulation using vision and touch, with implementation details provided in \cref{sec:implementation}. The proposed approach is empirically validated in \cref{sec:experiments} and concluded in \cref{sec:conclusion}.

\section{The \method Framework}\label{sec:method}

In this section, we present \method, a systematic framework for manipulating articulated objects by integrating vision and touch. We first introduce the contact-regulation methods that motivate our framework in \cref{sec:method-background}. These methods require a stable grasp and a coarse direction estimate as initialization. To address these requirements, we formulate the problem as a \acf{map} estimation task, decomposed into two tractable components in \cref{sec:method-problem_formulation}. Finally, we describe our approach for estimating a distribution over coarse motion directions without constraining the solution to specific articulation types in \cref{sec:method-vision}.

\subsection{Background: Contact-Regulating Methods}\label{sec:method-background}

Recent advances in articulated object manipulation demonstrate that kinematic priors are not strictly necessary if the robot regulates contact through tactile sensing~\cite{zhao2024tac,zhao2025tacman}. Given a coarse interaction direction, these methods iteratively adjust the end-effector pose by a transformation \(T_\Delta \in \mathrm{SE}(3)\) such that the resulting contact returns to a stable state. Formally, the update is computed as
\begin{equation}
    T_\Delta = \argmin_{T_\Delta \in \mathrm{SE}(3)} f(\mathcal{C}_0, \mathcal{C}_{t+1}),
    \label{eq:tactile_control}
\end{equation}
where \(\mathcal{C}_0\) denotes the reference contact, \(\mathcal{C}_{t+1}\) the contact after applying \(T_\Delta\), and \(f(\cdot,\cdot)\) a metric measuring their difference. By maintaining contact stability rather than tracking kinematic models, this formulation naturally handles objects with unknown or imprecisely estimated kinematics.

This kinematic-invariant property is precisely what enables reliable manipulation across diverse objects: vision modules need not recover error-prone hidden kinematics. However, successful execution requires two (i) a proper grasp that establishes stable contact and (ii) a coarse interaction direction to trigger the controller. Our framework addresses these requirements through principled visual inference.

\begin{figure}[b!]
    \centering
    \includegraphics[width=\linewidth]{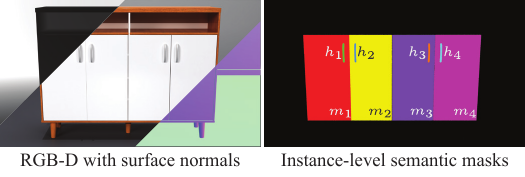}
    \caption{\textbf{Inputs to the vision module of \method.} The vision module of \method processes RGB-D data from an RGB-D sensor, surface normals computed from the depth map (visualized as a normal map), and instance-level semantic masks identifying holdable and movable parts. This representation accommodates objects with multiple interactable components. \textit{Note:} Holdable masks are subsets of their associated movable masks; regions appear overlapped in the visualization.}
    \label{fig:input}
\end{figure}

\subsection{Problem Formulation}\label{sec:method-problem_formulation}

Contact-regulating methods assume the availability of an initial stable grasp and a coarse interaction direction. Given visual observation $\mathcal{V}$, our goal is to recover these prerequisites by estimating:
\begin{itemize}[leftmargin=*,noitemsep,nolistsep]
    \item A parallel-gripper grasp \(G \in \mathrm{SE}(3) \times \mathbb{R}\), where the \(\mathrm{SE}(3)\) component specifies the gripper pose and the scalar encodes the gripper width.
    \item An interaction direction \(\boldsymbol{d} \in\mathbb{S}^2\), representing a unit vector on the 2-sphere.
\end{itemize}
Together, \((G, \boldsymbol{d})\) provide the initialization required for contact-regulation control.

In our setting, the visual observation consists of visually observable points \(\mathcal{V} = \{P_i\mid i = 1, \ldots, n\}\), where each point \(P_i\) is represented by:
\begin{equation}
    P = (\boldsymbol{p}, \boldsymbol{c}, \boldsymbol{n}, m, h).
    \label{eq:visual_input}
\end{equation}
As illustrated in \cref{fig:input}, \(\boldsymbol{p} \in \mathbb{R}^3\) denotes the 3D position in the camera frame, and \(\boldsymbol{c} \in [0, 255]^3\) represents RGB color. The surface normal \(\boldsymbol{n}\in\mathbb{S}^2\) provides geometric constraints that guide direction estimation beyond random guessing---a hypothesis we validate experimentally. The label \(m \in \mathbb{N}\) specifies whether the point is movable (\(m > 0\)) or fixed (\(m = 0\)), with different positive values corresponding to distinct movable parts within a single object. Similarly, \(h \in \mathbb{N}\) indicates whether the point provides a viable holdable location (\(h > 0\)) associated with a specific movable part. While position and color are obtained directly from depth sensing, the remaining attributes are inferred from them, as detailed in \cref{sec:implementation-segmentation}.

Formally, we seek to obtain:
\begin{equation}
    G^*, \boldsymbol{d}^* = \argmax_{G, \boldsymbol{d}} p(G, \boldsymbol{d}\mid\mathcal{V}),
    \label{eq:problem_formulation}
\end{equation}
where \(p(\cdot)\) represents a \ac{pdf}.

\begin{figure}[t!]
    \centering
    \includegraphics[width=\linewidth]{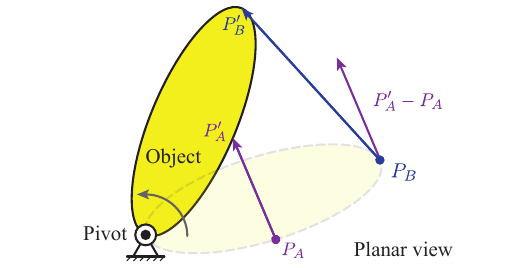}
    \caption{\textbf{Coupling between grasp point and interaction direction.} The interaction direction depends on the selected grasp point, even when the same rigid transformation is applied. Different point selections yield different directions under identical transformations.}
    \label{fig:coupling}
\end{figure}

Directly modeling the joint density \(p(G, \boldsymbol{d}\mid\mathcal{V})\) is challenging, yet treating \(G\) and \(\boldsymbol{d}\) as conditionally independent is not justified. As illustrated in \cref{fig:coupling}, the interaction direction depends on the grasp point \(\boldsymbol{g}\in\mathbb{R}^3\) determined by \(G\): even under the same rigid transformation, different grasp locations yield different directions. 

We make the problem tractable by modeling the rigid transformation \(T=[R\in\mathrm{SO(3)}\mid\boldsymbol{t}\in\mathbb{R}^3]\in\mathrm{SE(3)}\), which is independent of the specific grasping point. We then recover the interaction direction from \(T\) and point position \(\boldsymbol{p}\) via:
\begin{equation}
    \boldsymbol{d} = \frac{(R-I)\boldsymbol{p}+\boldsymbol{t}}{\|(R-I)\boldsymbol{p}+\boldsymbol{t}\|_2},
    \label{eq:d}
\end{equation}
where \(I\) is the \(3\times 3\) identity matrix. Then \cref{eq:problem_formulation} can be reformulated as:
\begin{align}
    G^*, T^* &= \argmax_{G, T} p(G, T\mid\mathcal{V})\\
             &= \underbracket{\argmax_{G} p(G\mid\mathcal{V})}_{\text{grasp}}\underbracket{\argmax_{T} p(T\mid\mathcal{V})}_{\text{direction}},
    \label{eq:problem_reformulation}
\end{align}
where \(p(G\mid\mathcal{V})\) and \(p(T\mid\mathcal{V})\) separately model grasp selection and transformation estimation. Since parallel-jaw grasping is well-studied and does not affect \(T\) estimation, we defer implementation details to \cref{sec:implementation-grasp}. The remainder of this section focuses on estimating the transformation distribution \(p(T\mid\mathcal{V})\), which determines the interaction direction.

\subsection{Vision-Based Direction Estimation}\label{sec:method-vision}

We detail our method for estimating the rigid transformation of a movable part from visual inputs. Unlike prior methods restricted to specific joint types such as revolute or prismatic joints, our approach makes no such assumption. Real-world articulated objects often deviate from these idealized models~\cite{zhao2024tac}, and although current datasets underrepresent such complexity, our method is designed to accommodate it.

Without assuming a predefined kinematic structure, we adopt a numerical approach to infer the rigid transformation. We introduce small perturbations to the movable part and analyze the resulting displacement patterns of associated points \(\boldsymbol{p}_i\) between consecutive frames. Each point acquires a displacement vector \(\boldsymbol{q}_i\in \mathbb{R}^3\) determined by \(T\):
\begin{equation}
    \boldsymbol{q}_i = T
    \begin{bmatrix}
        \boldsymbol{p}_i \\ 1
    \end{bmatrix}
    - \boldsymbol{p}_i.
    \label{eq:T}
\end{equation}
With sufficient point-displacement pairs \((\boldsymbol{p}_i, \boldsymbol{q}_i)\), we efficiently solve for \(T\) using the Kabsch algorithm~\cite{kabsch1976solution}.

Under the rigid-body assumption, every sub-part within the movable component undergoes the same transformation. Evaluating \cref{eq:T} with different point combinations therefore provides insight into the conditional probability distribution \(p(T\mid\mathcal{V})\). In an idealized scenario with perfect observations and strictly rigid motion, this distribution would collapse to a Dirac delta at the true transformation. Real-world conditions---noise, partial visibility, object complexity---introduce ambiguities that yield multiple plausible motion directions. This approach thus captures and represents uncertainties inherent in the vision-based model \(p(T\mid\mathcal{V})\).

With grasp point \(\boldsymbol{g}\) chosen to maximize the first term in \cref{eq:problem_reformulation}, we map each candidate transformation \(T\) to its corresponding interaction direction \(\boldsymbol{d}\) deterministically via \cref{eq:d}. This mapping induces a distribution over directions \(p(\boldsymbol{d}\mid\mathcal{V})\) from the underlying \(p(T\mid\mathcal{V})\). To model this distribution on the unit sphere \(\mathbb{S}^2\), we fit a \ac{vmf} distribution to the sampled directions \(\{\boldsymbol{d}_i\}_{i=1}^{n}\). The \ac{vmf} distribution is formulated as:
\begin{equation}
    p(\boldsymbol{d}\mid\mathcal{V}) = \frac{1}{c(\kappa, \boldsymbol{\mu})}
    \exp{\left(\kappa\boldsymbol{\mu}^\transpose \boldsymbol{d}\right)},\quad \boldsymbol{d}\in\mathbb{S}^2.
    \label{eq:vmf}
\end{equation}

Analogous to a Gaussian distribution in Euclidean space, the \ac{vmf} distribution employs two parameters: a mean direction \(\boldsymbol{\mu} \in \mathbb{S}^2\) specifying the central location, and a concentration parameter \(\kappa \in \mathbb{R}_{>0}\) controlling how tightly the distribution clusters around \(\boldsymbol{\mu}\). The normalizing constant \(c(\kappa, \boldsymbol{\mu})\) ensures that \(p(\boldsymbol{d}\mid\mathcal{V})\) integrates to one over \(\mathbb{S}^2\). 

Since the normalizing constant and \(\kappa\) do not affect the maximizer, we obtain:
\begin{equation}
    \argmax_{\boldsymbol{d}\in\mathbb{S}^2} p(\boldsymbol{d}\mid\mathcal{V}) = \argmax_{\boldsymbol{d}\in\mathbb{S}^2}
    \exp{\left(\boldsymbol{\mu}^\transpose \boldsymbol{d}\right)}.
\end{equation}

The density is maximized when \(\boldsymbol{d}\) aligns with \(\boldsymbol{\mu}\). We estimate \(\boldsymbol{\mu}\) by computing the Fréchet mean of sampled directions under the geodesic metric (arc length) on the sphere, yielding an unbiased estimator:
\begin{equation}
    \boldsymbol{d}^* = \hat{\boldsymbol{\mu}} = \argmin_{\boldsymbol{\mu}\in\mathbb{S}^2}{\sum_{i=1}^{n}{\left|\arccos{\left(\boldsymbol{\mu}^\transpose\boldsymbol{d}_i\right)}\right|^2}}.
    \label{eq:frechet_mean}
\end{equation}

\begin{figure}[t!]
    \centering
    \includegraphics[width=\linewidth]{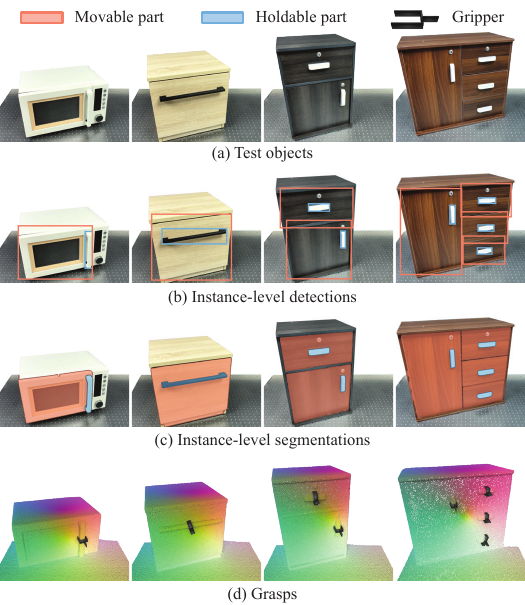}
    \caption{\textbf{Real-world articulated objects and processing pipeline.} (a) We evaluate \method on real-world objects spanning diverse configurations: prismatic to revolute joints, and single-part to multi-part structures. (b) Our trained detector reliably identifies movable and holdable parts, even in complex multi-part cases. (c) These detections provide prompts for the segmentation model, enabling fine-grained part segmentation. (d) Based on segmented parts, suitable grasps are generated at grasping points \(\boldsymbol{g}\). These results provide the necessary information for inferring interaction directions.}
    \label{fig:objects_rw}
\end{figure}

\section{Implementation}\label{sec:implementation}

In this section, we describe the implementation details of \method. We first introduce the dataset in \cref{sec:implementation-data}, which enables detection of movable and holdable parts in \cref{sec:implementation-segmentation}. We then explain how to leverage sampling-based models for stable grasping, followed by learning-based acquisition of point displacements using the established dataset in \cref{sec:implementation-vision}, which is critical for recovering interaction directions. Finally, we present the tactile control policy in \cref{sec:implementation-policy}.

\subsection{Dataset Preparation}\label{sec:implementation-data}

We construct a dataset to support learning-based extraction of movable and holdable features and direction estimation. We select 385 articulated objects spanning eight categories from the PartNet-Mobility dataset~\cite{xiang2020sapien} and import them into the SAPIEN simulator, rendering them in ray tracing mode from up to 72 viewpoints. This process captures the color and positional information defined in \cref{eq:visual_input}. Surface normals are estimated by computing the cross product of vectors formed from each point and its neighbors to the right and below in image space. Movable and holdable instance labels \(m\) and \(h\) are obtained from GAPartNet annotations~\cite{geng2023gapartnet}. 

The dataset is divided at the category level: microwaves, refrigerators, storage furniture, and trash cans are assigned to the training set, while dishwashers, doors, ovens, and tables are reserved for testing. Within the training portion, we split data into training and validation subsets using an 8:2 ratio, yielding \num{39524} training samples, \num{9881} validation samples, and \num{5836} test samples.

\begin{figure}[t!]
    \centering
    \includegraphics[width=\linewidth]{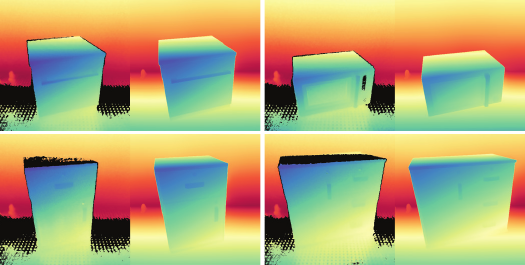}
    \caption{\textbf{Depth refinement using foundation models.} We leverage a depth foundation model~\cite{yang2024depth} to refine raw depth measurements from the image sensor. Left: raw depth. Right: refined depth. Both visualizations use the same colorbar range for comparability.}
    \label{fig:depth}
\end{figure}

To evaluate performance beyond simulation, we collect four real-world examples, each captured from five viewpoints using a Femto Bolt RGB-D sensor. One view is illustrated in \cref{fig:objects_rw}(a); additional results appear in the supplementary video. These examples capture real-world diversity, including objects with single and multiple movable parts, and are reserved strictly for testing~\cite{yang2024depth}. To improve depth quality, we first estimate a relative depth map using a depth foundation model~\cite{yang2024depth}. Since this estimate lacks an absolute scale, we recover the correct scale by fitting a linear model between estimated disparities and ground-truth sensor measurements using RANSAC for robustness. The enhancement is illustrated in \cref{fig:depth}.

\subsection{Movable and Holdable Part Segmentation}\label{sec:implementation-segmentation}

Using the prepared data from \cref{sec:implementation-data}, we derive the movable and holdable masks defined in \cref{eq:visual_input}, which serve as key inputs to our vision module. We train an object detector with a DINOv3 backbone and transformer-based head to detect movable and holdable parts~\cite{simeoni2025dinov3,carion2020end}. The model is trained using the AdamW optimizer on a single H100 GPU with batch size 2 and learning rate \(6\times10^{-6}\). Following the protocol suggested by Lin \etal~\cite{lin2014microsoft}, we report mean Average Precision across IoU thresholds from 0.50 to 0.95 (mAP@[0.50:0.95]). The model attains 0.86 mAP on the test set; detailed breakdowns appear in \cref{tab:detector}. Since mAP above 0.6 in multi-class settings is typically considered practically useful~\cite{lin2014microsoft} and detection is not our primary contribution, we provide the model and checkpoints in code rather than extensive baseline comparisons.

\begin{table}[ht!]
    \centering
    \small
    \caption{\textbf{Detection performance on the test set.}}
    \label{tab:detector}
    \begin{tabularx}{\linewidth}{*{6}{>{\centering\arraybackslash}X}}
        \toprule
        \textbf{mAP} & \textbf{AP(50)} & \textbf{AP(75)} & \textbf{AP(S)} & \textbf{AP(M)} & \textbf{AP(L)}\\
        \midrule
        0.86 & 0.97 & 0.94 & 0.66 & 0.86 & 0.94 \\
        \bottomrule
    \end{tabularx}
\end{table}

Detector outputs are passed to SAM2~\cite{ravi2024sam} to produce final movable and holdable masks on an RTX 3090 GPU. We associate each holdable part with its corresponding movable part by selecting the pair whose mask intersection has the largest area. Real-world results are presented in \cref{fig:objects_rw}(b)--(c) for illustration.

\subsection{Grasp Selection}\label{sec:implementation-grasp}

With movable and holdable masks defined, we establish a stable grasp on the handle. Recent advances demonstrate the effectiveness of parallel grippers for object grasping, even in cluttered environments~\cite{mahler2019learning,sundermeyer2021contact,fang2023anygrasp}. The handle-grasping problem is largely simplified in our setting. We adopt a sampling-based method similar to Ten \etal~\cite{ten2017grasp}, restricting the grasp region to the holdable area. The grasping point \(\boldsymbol{g}\) is defined as the centroid of this region, which determines the gripper translation. We sample gripper rotations to identify one yielding a collision-free grasp with minimal gripper width. Considering the symmetry of the parallel gripper, we select the pose closest to the robot's home position~\cite{zhao2025b}. Qualitative examples appear in \cref{fig:objects_rw}(d).

\begin{figure}[t!]
    \centering
    \includegraphics[width=\linewidth]{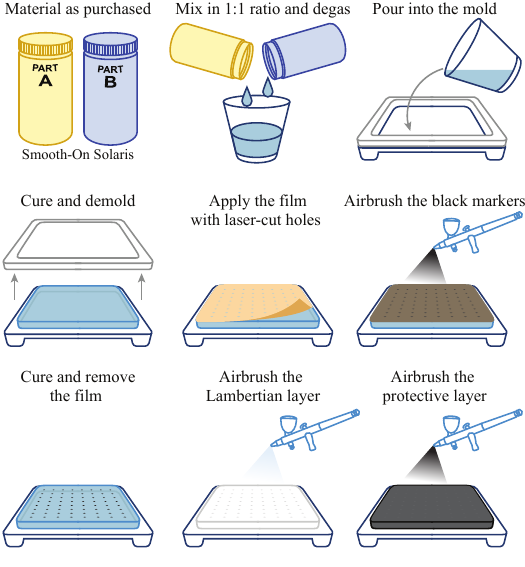}
    \caption{\textbf{Fabrication process for GelSight-style tactile sensor elastomer.} We use Smooth-On Solaris silicone as the base elastomer. Marker placement is standardized using a laser-cut stencil to ensure uniform spacing and geometry. The Lambertian coating and protective topcoat are applied via airbrush.}
    \label{fig:tactile}
\end{figure}

\subsection{Vision-Based Displacement Estimation}\label{sec:implementation-vision}

We estimate the displacement flow from visual inputs defined in \cref{eq:T} using a neural network based on PointNet++~\cite{qi2017pointnet++}. The network takes point coordinates as input and augments them with surface normals in the movable region, along with movable masks as additional features.

Training uses the loss:
\begin{equation}
    \mathcal{L} = \underbracket{\frac{1}{n}\sum_{i=1}^n \frac{\|\hat{\boldsymbol{q}}_i - \boldsymbol{q}_i\|_1}{\|\boldsymbol{q}_i\|_1}}_{\text{magnitude}}+\underbracket{\frac{1}{n}\sum_{i=1}^n\left(1-\frac{\hat{\boldsymbol{q}}_i^{\transpose}\boldsymbol{q}_i}{\|\hat{\boldsymbol{q}}_i\|_2\|\boldsymbol{q}_i\|_2}\right)}_{\text{direction}},
    \label{eq:loss}
\end{equation}
where \(n\) is the number of points and \(\hat{\boldsymbol{q}}_i\) is the network's estimate. The first term penalizes magnitude error using relative \(\ell_1\) loss, which stabilizes optimization across a wide dynamic range and drives equal absolute errors toward zero regardless of scale. This is important because small displacements arise both outside masks and within masked regions near rotation axes. The second term aligns predicted and target directions via cosine similarity, ensuring accurate orientation even when magnitudes are small. The model is trained using the AdamW optimizer on a single H100 GPU with a batch size of 32 and a learning rate of \(1\times10^{-3}\). Inference is performed on an RTX 3090 GPU for all experiments.

\begin{figure}[t!]
    \centering
    \includegraphics[width=\linewidth]{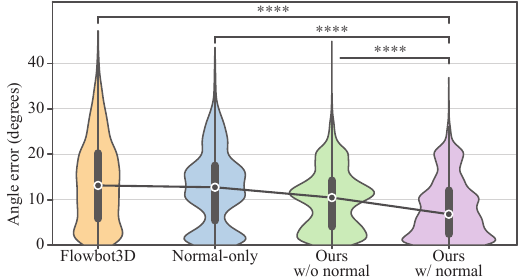}
    \caption{\textbf{Quantitative results of direction estimation on unseen object categories.} Prediction errors from four methods over \num{5836} test samples drawn from categories not seen during training. \method, which uses surface normals as an inductive bias, achieves significant performance gains over baselines. The violin plots show error distributions: the outer shape is the kernel density estimate (KDE); the white dot is the median; the thick bar denotes the interquartile range (IQR); and the whiskers extend to \(1.5\times\) IQR beyond the quartiles. Note: **** indicates \(p < 0.0001\).}
    \label{fig:direction}
\end{figure}

\begin{table}[t!]
    \centering
    \small
    \caption{\textbf{Quantitative comparison of interaction direction estimation methods on the test set.} Results are reported as the mean and standard deviation of angle errors (in degrees).}
    \label{tab:direction}
    \begin{tabular}{@{}*{4}{S[table-format=2.2\pm1.2,separate-uncertainty=true,retain-zero-uncertainty=true]}@{}}
        \toprule
        \text{Flowbot3D} & \text{Normal-only} & \text{Ours w/o normal} & \text{Ours w/ normal}\\
        \midrule
        13.92\pm9.45 & 12.66\pm8.25 & 10.10\pm6.57 & \bfseries8.13\pm6.54\\
        \bottomrule
    \end{tabular}
\end{table}

\subsection{Tactile Manipulation Policy}\label{sec:implementation-policy}

Following initialization by the estimated grasp and interaction direction, all subsequent manipulation is governed by a tactile-based policy. We employ a tactile controller to manipulate articulated objects, building on the work of Zhao \etal~\cite{zhao2024tac,zhao2025tacman}, which utilizes a GelSight-style tactile sensor~\cite{yuan2017gelsight} to provide contact feedback. This approach extracts tactile features from the positions of activated markers—defined as those whose normal deformation exceeds a predefined threshold. By tracking marker-wise position changes, the controller computes pose updates (\cref{eq:tactile_control}) via a point registration algorithm operating at \SI{50}{\hertz}. For complete algorithmic details, we refer readers to the original work due to space constraints.

While GelSight-style sensors are widely adopted and their mechanical design and calibration are well documented~\cite{yuan2017gelsight,li2024minitac,zhao2025embedding,li2025taccel}, fabrication of the core component---the elastomer with Lambertian coating---appears to remain lab-specific. To improve reproducibility, we detail one practical fabrication procedure used in this study in \cref{fig:tactile}. The airbrushable silicone pigment is prepared by mixing silicone pigment with Smooth-On Psycho Paint (a platinum-silicone paint base) and thinning the mixture using Smooth-On NOVOCS Matte solvent. This enables uniform spray application and consistent elastomer finishes suitable for tactile imaging.

\section{Experiments}\label{sec:experiments}

This section evaluates \method through comprehensive experiments. We begin with large-scale tests on synthetic objects in \cref{sec:experiments-simulation} to assess generalization across unseen categories. We then validate \method in the real world (\cref{sec:experiments-real_world}), demonstrating the complete pipeline for manipulating unknown articulated objects via vision and touch.

\begin{figure}[t!]
    \centering
    \includegraphics[width=\linewidth]{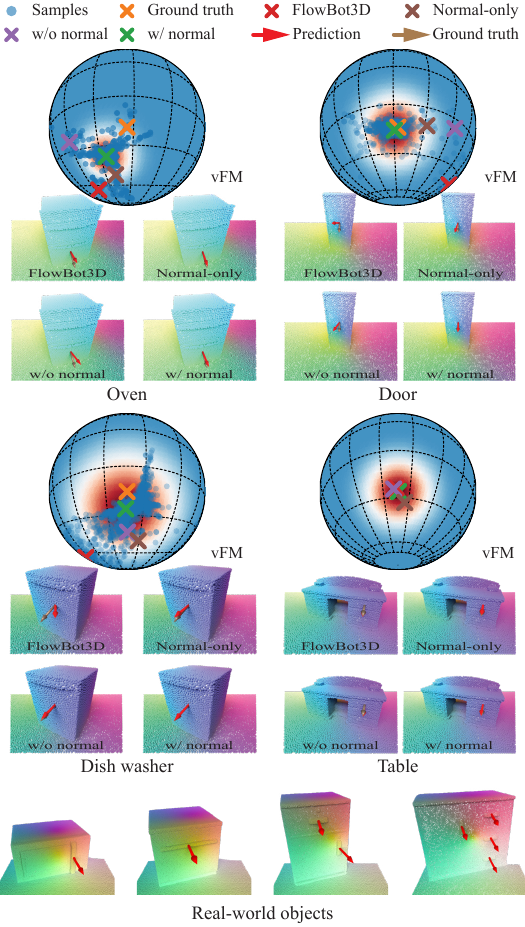}
    \caption{\textbf{Qualitative results of direction estimation on unseen object categories.} We illustrate the approach using four representative objects, one from each test category. For each object, we show the obtained samples, the fitted \ac{vmf} distribution, the ground truth, and predictions from the three baseline methods. By fitting the distribution and incorporating surface normals as an inductive bias, \method demonstrates greater robustness to high uncertainty when encountering previously unseen objects. The bottom row presents results on real-world examples using the grasping points shown in \cref{fig:objects_rw}(d), demonstrating successful transfer from simulation to real-world settings.}
    \label{fig:sim}
\end{figure}

\begin{figure}[t!]
    \centering
    \includegraphics[width=\linewidth]{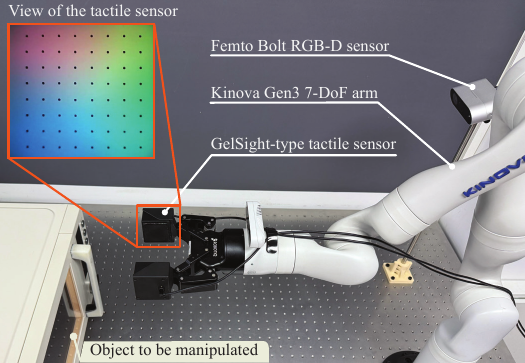}
    \caption{\textbf{Experimental platform for real-world validation.}}
    \label{fig:exp_set}
\end{figure}

\begin{figure}[bh!]
    \centering
    \includegraphics[width=\linewidth]{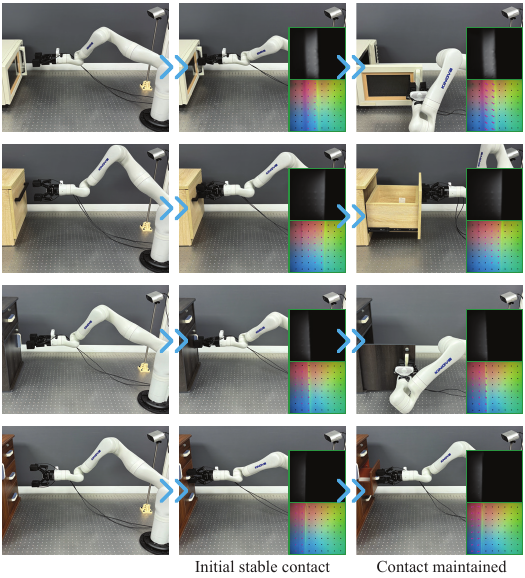}
    \caption{\textbf{Real-world validation of \method.} Leveraging visual cues, the robot automatically establishes stable contact with the handle of the articulated object. Following the estimated interaction direction, the low-level, tactile-informed controller reliably completes the manipulation.}
    \label{fig:real_world}
\end{figure}

\subsection{Simulation Studies}\label{sec:experiments-simulation}

We evaluate interaction direction estimation using \num{5836} test samples from categories unseen during training, as introduced in \cref{sec:implementation-data}. This setup allows us to assess generalization to previously unknown articulated objects. We compare \method, which leverages surface normals as an important inductive bias, against three baselines:
\begin{itemize}[leftmargin=*,noitemsep,nolistsep]
    \item \textbf{FlowBot3D:} A recent method for articulated object manipulation that employs point-displacement modeling similar to ours~\cite{ebner2025stein}. It selects the interaction direction that maximizes articulation movement without modeling the full direction distribution.
    \item \textbf{Normal-only:} A simple, learning-free baseline that computes the Fréchet mean (see \cref{eq:frechet_mean}) of surface normals within the moving region.
    \item \textbf{Without-normal:} An ablation that trains our model without surface-normal inputs while keeping all other components unchanged, isolating the contribution of this feature.
\end{itemize}

For fair comparison, we set the grasping point as the movable region's centroid for both the \method and the Without-normal baseline. For FlowBot3D and Normal-only, we translate predictions to grasping points for comparison.

Quantitative results are shown in \cref{fig:direction,tab:direction}, where prediction error is measured as the angle between predicted and ground-truth directions. All four methods achieve median errors around \SI{10}{\degree}, highlighting the challenge of recovering precise motion directions on unfamiliar geometries. FlowBot3D, without modeling the distribution of point displacements, shows greater sensitivity to unseen categories. The Normal-only baseline, despite its simplicity, achieves competitive performance. \method reduces uncertainty by modeling the distribution of fitted directions, and explicitly incorporating surface normals further improves performance. One-sided paired t-tests confirm statistically significant improvements over all three baselines (\(p < 0.0001\)). 

For qualitative illustration, \cref{fig:sim} shows four representative objects from the test categories, visualizing sample directions alongside the corresponding fitted \ac{vmf} distributions.

\subsection{Real-World Experiments}\label{sec:experiments-real_world}

To assess the gap between synthetic objects and real-world scenarios, we evaluate our model on physical objects captured in the real world, as shown in \cref{fig:objects_rw}(a). Using the selected grasping points in \cref{fig:objects_rw}(d), we present four representative examples in \cref{fig:sim}, demonstrating that \method generates plausible interaction direction estimates.

To further assess whether visual cues alone can drive complete manipulation of articulated objects, we implement the full pipeline in the real world, from vision-based high-level guidance to tactile-informed low-level control. We use a Kinova Gen3 7-DoF arm equipped with GelSight-type tactile sensors in place of its default gripper pads, as described in \cref{sec:implementation-policy}. The integrated system is illustrated in \cref{fig:exp_set}.

As shown in \cref{fig:real_world}, \method guides the robot to reliably establish valid grasps on real objects and follow the estimated interaction direction. By leveraging tactile feedback, the system adapts its motions in real time (\SI{50}{\hertz}), achieving consistent and robust manipulation across all articulated objects. The complete manipulation process and additional experimental results are provided in the supplementary materials and on our website.

\section{Conclusion and Future Work}\label{sec:conclusion}

We introduced \method, a framework for articulated object manipulation that leverages the complementary strengths of vision and touch. Rather than inferring precise but unreliable kinematics from vision alone, \method uses vision for coarse guidance---grasp proposals and interaction direction estimates---while relying on tactile feedback for robust execution. By incorporating surface normals as a geometric prior and modeling interaction directions with a \ac{vmf} distribution, \method generalizes to unseen objects and outperforms existing baselines. Our evaluations demonstrate that \method enables autonomous manipulation of diverse articulated objects without explicit kinematic models, highlighting the value of integrating visual guidance with tactile control.

\paragraph*{Interpretability through hierarchical design}

A key advantage of \method's hierarchical architecture---which separates visual intention from tactile execution---is its inherent interpretability. Unlike end-to-end policies that map pixels directly to actions, our system explicitly generates a coarse interaction direction $\boldsymbol{d}$ and grasp $G$ before contact is made. This intermediate representation serves as a communicable ``intention'' that could be exposed to human users in future iterations. For instance, augmented reality projections of the intended trajectory or verbal announcements (\eg, ``Opening cabinet'') prior to execution would foster safer, more predictable human-robot interaction and simplify debugging for safety certification in unstructured domestic environments.

\paragraph*{Extending to non-rigid and multi-modal scenarios}

Our current formulation leverages the Kabsch algorithm under a strict rigid-body assumption to estimate displacement. However, domestic objects often exhibit compliance or multi-stage articulation (\eg, flexible handles or nested joints). Future work will explore extending our displacement estimation to handle non-rigid deformations, potentially through deformable object tracking or sequential state estimation. Additionally, while our \ac{vmf}-based modeling captures directional uncertainty, objects with multiple distinct valid interaction directions (\eg, a lever that can toggle both up and down) may require multi-modal distribution modeling or conditioning on high-level user commands.

\paragraph*{Handling grasp failures}

Our method currently assumes the initial grasp remains stable throughout manipulation. To mitigate failures from grasp slippage, we plan to integrate tactile slip detection with dynamic re-grasping primitives, enabling the system to recover from execution errors and improve overall robustness.

\balance
\bibliographystyle{ieeetr}
\bibliography{reference_header,reference}

\end{document}